\documentclass[letterpaper, 10 pt, conference]{ieeeconf}  %

\IEEEoverridecommandlockouts                              %

\usepackage[backend=bibtex,
            hyperref=false,
            url=false,
            isbn=false,
            doi=false,
            backref=false,
            style=numeric-comp,
            sortcites,
            block=none]{biblatex}
            
\renewcommand{\bibfont}{\small}
\usepackage{multirow}

\usepackage{algpseudocode}
\usepackage{algorithmicx}
\usepackage[ruled]{algorithm}

\usepackage{color,xcolor}
\usepackage{epsfig}
\usepackage{graphicx}

\usepackage{adjustbox}
\usepackage{array}
\usepackage{booktabs}
\usepackage{colortbl}
\usepackage{float,wrapfig}
\usepackage{hhline}
\usepackage{multirow}
\usepackage{subcaption} %

\usepackage{bm}
\usepackage{nicefrac}
\usepackage{microtype}

\usepackage{changepage}
\usepackage{extramarks}
\usepackage{fancyhdr}
\usepackage{lastpage}
\usepackage{setspace}
\usepackage{soul}
\usepackage{xspace}

\usepackage{url}

\usepackage{enumerate}

\usepackage{titlesec}
\usepackage{makecell}

\usepackage{mathtools}
\usepackage{stmaryrd}
\usepackage{hyperref}
\hypersetup{
    colorlinks=true,
    linkcolor=blue,
    filecolor=magenta,      
    urlcolor=cyan,
    }

\usepackage{tabularx}

\newcolumntype{L}[1]{>{\raggedright\let\newline\\\arraybackslash\hspace{0pt}}m{#1}}
\newcolumntype{C}[1]{>{\centering\let\newline\\\arraybackslash\hspace{0pt}}m{#1}}
\newcolumntype{R}[1]{>{\raggedleft\let\newline\\\arraybackslash\hspace{0pt}}m{#1}}

\makeatletter
\DeclareRobustCommand\onedot{\futurelet\@let@token\@onedot}
\def\@onedot{\ifx\@let@token.\else.\null\fi\xspace}

\g@addto@macro\normalsize{%
  \setlength\abovedisplayskip{0pt}
  \setlength\belowdisplayskip{0pt}
  \setlength\abovedisplayshortskip{0pt}
  \setlength\belowdisplayshortskip{0pt}
}
\makeatother

\newcommand{\ours}{Local Neural Descriptor Field\xspace}
\newcommand{\ourshort}{L-NDF\xspace}

\newcommand{\sect}[1]{Section~\ref{#1}}

\newcommand{\ignore}[1]{}

\newcommand{\coord}{\mathbf{x}}
\newcommand{\pointcloud}{\mathbf{P}}

\newcommand{\pose}{\mathbf{T}}

\newcommand{\pointdescriptor}{z}

\newcommand{\sethree}{$\text{SE}(3)$}

\newcommand{\idealpose}{\pose^*}

\newcommand{\SE}[1]{\text{SE}(#1)}

\newcommand{\pointndf}{f}
\newcommand{\posendf}{F}

\newcommand{\posedescriptor}{\mathcal{Z}}

\newcommand{\demopointcloud}{\hat{\pointcloud}}

\newcommand{\demopose}{\hat{\pose}}

\newcommand{\probes}{\mathcal{X}}

\usepackage{color}
\definecolor{turquoise}{cmyk}{0.65,0,0.1,0.3}
\definecolor{purple}{rgb}{0.65,0,0.65}
\definecolor{dark_green}{rgb}{0, 0.5, 0}
\definecolor{orange}{rgb}{0.8, 0.6, 0.2}
\definecolor{red}{rgb}{0.8, 0.2, 0.2}
\definecolor{darkred}{rgb}{0.6, 0.1, 0.05}
\definecolor{blueish}{rgb}{0.0, 0.3, .6}
\definecolor{light_gray}{rgb}{0.7, 0.7, .7}
\definecolor{pink}{rgb}{1, 0, 1}
\definecolor{greyblue}{rgb}{0.25, 0.25, 1}

\definecolor{blueish}{rgb}{0.0, 0.3, .6}

\DeclareRobustCommand{\vsnote}[1]{\ifthenelse{\boolean{draft-mode}}{\textcolor{dark_green}{[VS: #1}]}{}}

\DeclareRobustCommand{\pa}[1]{\ifthenelse{\boolean{draft-mode}}{\textcolor{green}{\textbf{[PA: #1}]}}{}}
\DeclareRobustCommand{\asnote}[1]{\ifthenelse{\boolean{draft-mode}}{\textcolor{blue}{\textbf{AS: #1}}}{}}
\DeclareRobustCommand{\arnote}[1]{\ifthenelse{\boolean{draft-mode}}{\textcolor{green}{\textbf{AR: #1}}}{}}

\newcommand{\Fig}[1]{Fig.~\ref{fig:#1}}

\newcommand{\Table}[1]{Table~\ref{tbl:#1}}
\newcommand{\eq}[1]{(\ref{eq:#1})}

\usepackage{blindtext}

\renewcommand{\paragraph}[1]{\vspace{.1em}\noindent\textbf{#1}.}

\usepackage{amsmath,amsfonts,bm}

\def\eqref#1{equation~\ref{#1}}

\def\1{\bm{1}}

\DeclareMathAlphabet{\mathsfit}{\encodingdefault}{\sfdefault}{m}{sl}
\SetMathAlphabet{\mathsfit}{bold}{\encodingdefault}{\sfdefault}{bx}{n}

\DeclarePairedDelimiterX{\infdivx}[2]{(}{)}{%
  #1\;\delimsize|\delimsize|\;#2%
}

\title{\ours{}s:\\ Locally Conditioned Object Representations for Manipulation}

\author{Ethan Chun$^{1}$, Yilun Du$^{1}$, Anthony Simeonov$^{1}$, Tomas Lozano-Perez$^{1}$, Leslie Kaelbling$^{1}$ \\% <-this %
$^{1}$Computer Science and Artificial Intelligence Laboratory, MIT, USA 
}

\begin{document}

\twocolumn[{%
\renewcommand\twocolumn[1][]{#1}%
\maketitle
\begin{center}
\centering
\captionsetup{type=figure}

\includegraphics[width=\linewidth]{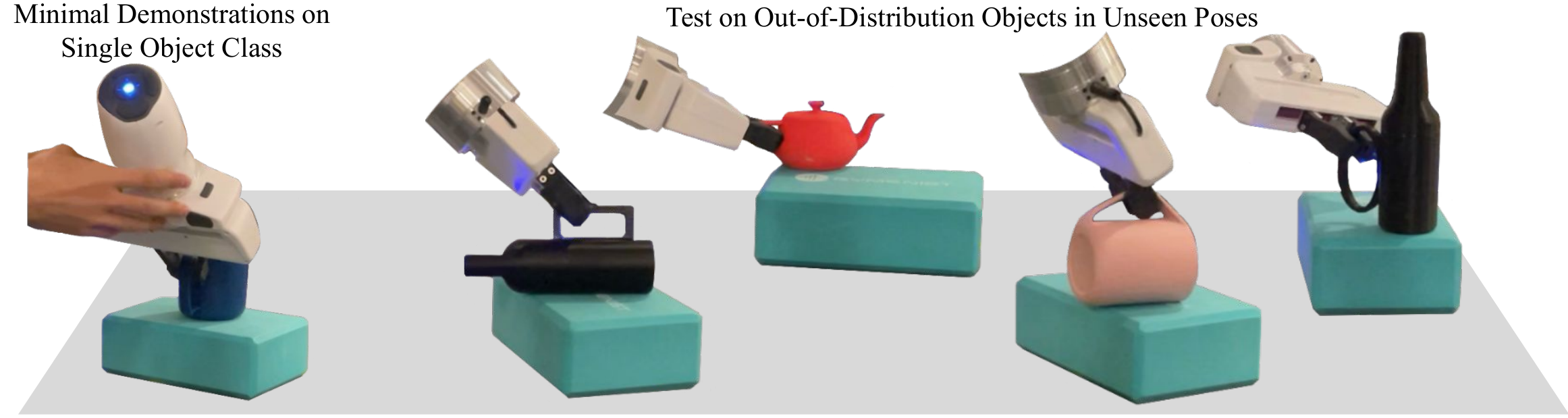}

\captionof{figure}{\small{Given minimal (5-10) real world demonstrations of grasping and picking up two different upright mugs, \ours can successfully grasp and pick up a set of geometrically distinct objects at arbitrary SE(3) poses.  }}

\label{fig:teaser}
\end{center}

}]

\global\csname @topnum\endcsname 0
\global\csname @botnum\endcsname 0

\thispagestyle{empty}
\pagestyle{empty}

\begin{abstract}
A robot operating in a household environment will see a wide range of unique and unfamiliar objects. While a system could train on many of these, it is infeasible to predict all the objects a robot will see.
In this paper, we present a method to generalize object manipulation skills acquired from a limited number of demonstrations, to novel objects from unseen shape categories. Our approach, Local Neural Descriptor Fields (L-NDF), utilizes neural descriptors defined on the local geometry of the object to effectively transfer manipulation demonstrations to novel objects at test time.  In doing so, we leverage the local geometry shared between objects to produce a more general manipulation framework. We illustrate the efficacy of our approach in manipulating novel objects in novel poses -- both in simulation and in the real world. 
Project website, videos, and code: \url{https://elchun.github.io/lndf/}. 
\footnotetext{Correspondence to: yilundu@mit.edu}

\end{abstract}

\section{Introduction}

A robot operating autonomously in an household environment will encounter a wide variety of unseen objects. While individual objects may be novel in shape, many can be decomposed into a set of previously seen constituent parts. Consider the novel objects illustrated in \Fig{teaser} -- while a bottle with a handle may be unseen, both bottles and mugs are individually known. Therefore, one may propose that a robot manipulate the novel object via skills learned on both bottles and mugs. In this paper, we investigate enabling such generalization using an imitation learning paradigm. In particular, we wish to construct a system which, when given a small set (5 - 10) of manipulation demonstrations on a single category of objects, can successfully execute this skill on novel objects types in arbitrary SE(3) orientations. 

To enable efficient learning, we build on the Neural Descriptor Fields (NDF) system~\cite{simeonov2021neural}. NDF assigns a dense descriptor to each point in a shape, with similar points across different objects in a given category assigned similar descriptors. Object manipulation may be generalized to novel objects in the same category by finding a corresponding set of dense descriptors in the novel object. A limitation of NDF, however, is that it relies on a single global latent to encode all geometric aspects of a shape in a given category. When given an object of a new category, this representation cannot capture the resultant geometry, preventing NDF from transferring object manipulation to new categories of objects.

We circumvent this problem by using a voxel grid of latents to locally capture the geometry and descriptors of a shape (see \Fig{model}); where each latent encodes a local spatial region. With this encoding scheme, descriptors of shapes in new categories can be more accurately encoded, as individual patches of the new shape correspond to patches from various categories of training object. We illustrate how this encoding enables generalization of object manipulation to new categories, referring to our approach as Local Neural Descriptor Fields (L-NDF).

An issue that arises when encoding descriptors locally is that descriptors of a portion of an object may change as the object is transformed. For example, the handle of a mug is represented with different voxel latents when it is translated and rotated. To ensure that descriptors are consistent across rigid object transformations, we propose a contrastive loss which explicitly enforces descriptor consistency when objects are transformed.

To transfer object manipulation demonstrations from one object to another, we must find corresponding sets of descriptors between the objects.  In NDFs, a global gradient optimization procedure is used to minimize descriptor distance.  
With L-NDF, a similar global optimization procedure is difficult to run, as descriptors of an object are only locally encoded -- lacking a consistent global direction in which descriptors are changing in a shape. To overcome this difficulty, we propose to initialize optimization across a diverse set of positions in a shape -- running local optimization to choose the descriptor with a minimal descriptor distance as our final, matching, descriptor.

We demonstrate that L-NDFs can be reliably used to generalize object manipulation to both novel objects and objects at unseen SE(3) poses. Given only (5-10) demonstration, our framework is able to manipulate novel objects (such as a tea cup or a bowl with a handle attached) in both simulation as well as on a real robot.
\section{Related Work}
\label{sec: related}

\subsection{Generalizable Manipulation}
Our work follows a long line of work on using imitation learning for manipulation. When object models are known, pose estimation may be used for manipulation~\cite{yoon2003real, zhu2014single, Schulman2016}. When the precise geometry of objects is unknown, template matching with coarse 3D primitives~\cite{miller2003automatic, harada2013probabilistic, LIS277_shapeBasedSkillTransfer_icra2021} or nonrigid registration~\cite{Schulman2016} can be used; but such methods still suffer when objects deviate substantially from templates. Recent work has explored more flexible representations for imitation learning, such as keypoint~\cite{gao2019kpam,manuelli2019kpam, gao2021kpam} or dense descriptors~\cite{florence2018dense,sundaresan2020learning,simeonov2021neural}. Most similar to our work -- DON~\cite{florence2018dense} and NDF explore 2D and 3D dense descriptors for object manipulation -- but both only demonstrate generalization within the same category of objects. In contrast, our approach enables object manipulation for novel categories of shapes at test time.

\vspace{-3pt}
\subsection{Neural Implicit Representations for Robotics}
Neural implicit representations~\cite{park2019deepsdf,mescheder2019occupancy} have emerged as a promising representation of 3D geometry in robotics. Different works have explored how implicit representations may be used in navigation~\cite{adamkiewicz2022vision}, localization~\cite{yen2020inerf,moreau2022lens,fu2022robust}, SLAM~\cite{Sucar:etal:ICCV2021,zhu2022nice,Ortiz:etal:iSDF2022}, and manipulation ~\cite{simeonov2021neural,jiang2021synergies,IchnowskiAvigal2021DexNeRF,yen2022nerfsupervision,simeonov2022rndf,ryu2022equivariant}. 
In the context of manipulation, ~\cite{IchnowskiAvigal2021DexNeRF, yen2022nerfsupervision} utilize NeRF as an approach to extract the underlying 3D geometry of a scene. In contrast, ~\cite{simeonov2021neural, simeonov2022rndf, ryu2022equivariant} build on the Neural Descriptor Field framework for learning manipulation skills, where underlying high-dimensional neural descriptors are used to transfer and generalize demonstrations. Our work extends NDFs to work with locally conditioned implicit representations. 

\section{Background: Manipulation with Neural Descriptor Fields}
\label{sect:background}

A Neural Descriptor Field (NDF)~\cite{simeonov2021neural} encodes the shape of an object using a function $\pointndf$ that maps a 3D point $\coord \in \mathbb{R}^{3}$ and an partial object point cloud $\pointcloud \in \mathbb{R}^{3\times N}$ to a spatial descriptor in $\mathbb{R}^{d}$: 
\begin{align}
    \pointndf(\coord | \pointcloud): \mathbb{R}^3 \times \mathbb{R}^{3 \times N} \to \mathbb{R}^d.
\end{align}
NDFs are also trained to learn correspondence over objects in the same category, so that points near similar geometric features of different instances (e.g., a point near the neck of two different bottles) are mapped to similar descriptor values.

NDFs can be generalized to assign descriptors to full \sethree{} poses, rather than individual points.
This is achieved by concatenating the descriptors of the individual points in a \emph{rigid set} of query points $\probes \in \mathbb{R}^{3\times N_q}$, i.e., a set of three or more non-collinear points $\coord_i, ~i=1...N_q$, that are constrained to transform together rigidly.
This construction allows NDFs to represent an \sethree{} pose $\pose$ via its action on $\probes$, i.e., via the points of the \emph{transformed query point cloud} $\pose \probes$: %
\begin{equation}
\posedescriptor = \posendf(\pose | \pointcloud) = \bigoplus_{\coord_i \in \probes} \pointndf(\pose \coord_i |\pointcloud)
\label{eq:pose_enc}
\end{equation}
Thus, $\posendf$ maps a point cloud $\pointcloud$ and an \sethree{} pose $\pose$ to a category-level pose descriptor $\posedescriptor \in \mathbb{R}^{d \times N_{q}}$.

\paragraph{Few-Shot Manipulation Learning with NDFs} Next, we discuss how to leverage NDF for few-shot learning of object manipulation skills. Consider a set of $K$ demonstrations, $\{\mathcal{D}_{i}\}_{i=1}^{K}$, where each demonstration, $\mathcal{D}_i=(\pointcloud^{i}, \pose_{pick}^i, \pose_{place}^i)$ consists of a object $\pointcloud^{i}$, and two poses: the end-effector pose before grasping, $\pose_{pick}^i$, and the relative pose of the placement surface $\pose^{i}_{place}$. We define a set of query points $\probes_{pick}$ and $\probes_{place}$ to represent the gripper and placement surface, respectively. We then utilize ~\eq{pose_enc} to encode each pose $\pose^i_{*}$ into its vector of descriptors $\mathcal{Z}^i_{*}$, conditional on the respective object point cloud $\pointcloud^i$, obtaining a set of spatial descriptor tuples $\{(\mathcal{Z}^i_{pick}, \mathcal{Z}^i_{rel})\}_{i=1}^{K}$. The set of descriptors is averaged over the $K$ demonstrations to obtain \emph{single} pick and place descriptors $\bar{\mathcal{Z}}_{pick}$ and $\bar{\mathcal{Z}}_{rel}$

When a new object is placed in the scene at test time, we obtain a point cloud $\pointcloud^{test}$ and leverage~\eq{energy} to recover $\pose^{test}_{pick}$ and $\pose^{test}_{rel}$ by minimizing the distance to spatial descriptors $\bar{\mathcal{Z}}_{pick}$ and $\bar{\mathcal{Z}}_{rel}$. 
\begin{align}
\bar \pose = \underset{\pose}{\text{argmin}} \|
 \posendf(\pose | \pointcloud) -
\posendf(\demopose | \demopointcloud)  \|
\label{eq:energy}
\end{align}
We rely on off-the-shelf inverse kinematics and motion planning algorithms to execute the final predicted poses.

\section{Local Neural Descriptor Fields}

Given a set of $K$, single object class, pick and place demonstrations, $\{\mathcal{D}_{i}\}_{i=1}^{K}$, where each demonstration, $\mathcal{D}_i=(\pointcloud^{i}, \pose_{pick}^i, \pose_{place}^i)$, consists of a partial object point cloud $\pointcloud^{i}$, end-effector pick pose $\pose_{pick}^i$ and place pose $\pose_{place}^i$, 
we are interested in generalizing the tasks to a set of new objects $\pointcloud'$ from unseen object classes. 
To solve this problem, we develop an approach using locally defined descriptors and propose suitable modifications of the NDF pipeline (\sect{sect:background}) to utilize such descriptors.

In particular, in \sect{sect:LNDF}, we introduce \ours{s} and illustrate how they may be used to locally encode the geometry of objects. 
In \sect{sect:equiVar}, we discuss how we may build SE(3)  equivariance into the underlying descriptor of \ourshort.  
Finally, in \sect{sect:poseOpt}, we discuss how to modify the underlying optimization procedure to allow us to search for an ideal pose within the local descriptor field landscape.  

\subsection{Local Descriptor Fields}
\label{sect:LNDF}

\begin{figure}[t]
\centering

\includegraphics[width=\columnwidth]{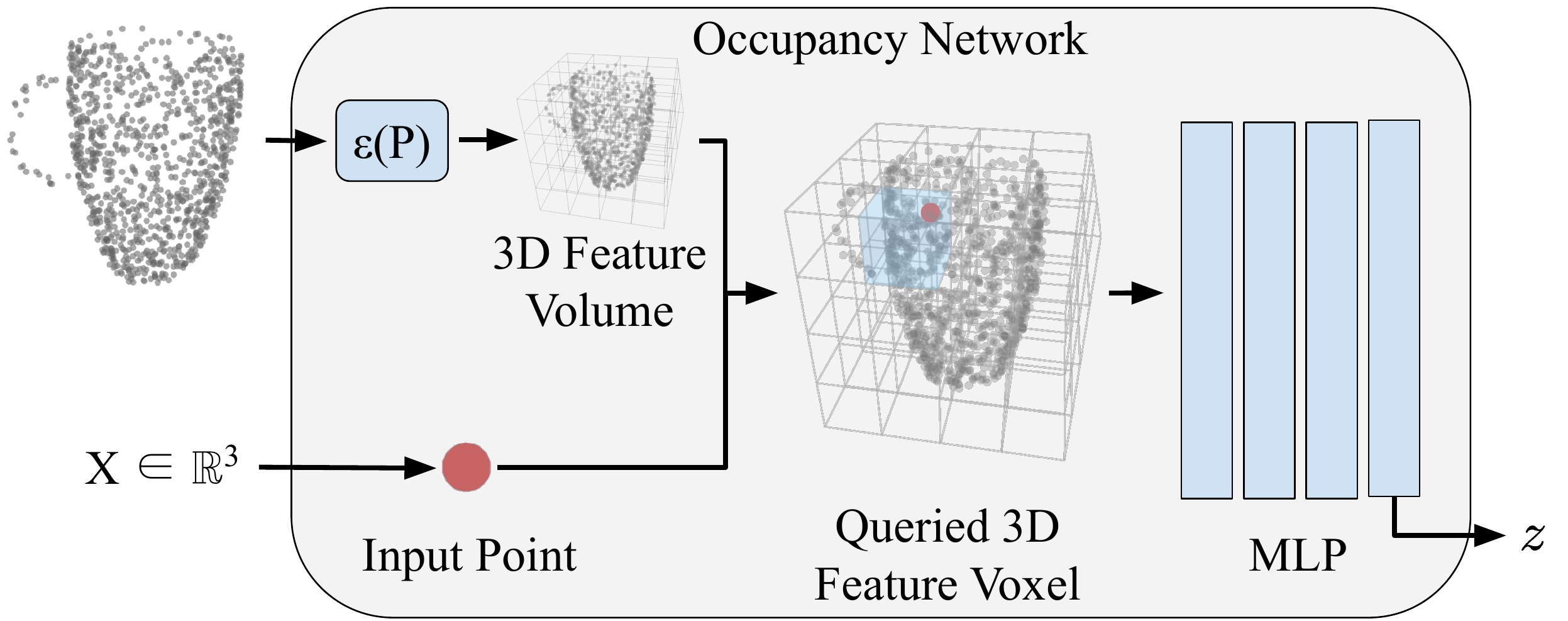}
\vspace*{-3.9mm}
\caption{\small
\textbf{\ours Architecture} -- A L-NDF takes any coordinate in 3D space, $\coord$, and a conditioning point cloud $\pointcloud$.  It then uses an encoder $\epsilon(\pointcloud)$ to encode $\pointcloud$ into a 3D feature volume from which the voxel containing $\coord$ is queried.  These feature are passed into an MLP decoder where the activations of the decoder's final layer are extracted to create the spatial descriptor, $\pointdescriptor$.
}
\label{fig:model}
\vspace{-15pt}
\end{figure}

A global NDF model cannot generalize effectively to new categories of objects.  To solve this problem, we use local descriptor fields for objects: each element of a voxel grid contains a latent vector representation of the object's local shape near that voxel.

In L-NDF, we use a convolutional occupancy network encoder ~\cite{peng2020convolutional}, $\epsilon(\pointcloud)$, to encode a partial point cloud $\pointcloud$ into a voxel grid of latents (illustrated in \Fig{model}). When querying a particular point, $\coord$, the corresponding voxel from the latent feature, $\epsilon(\pointcloud)$, is retrieved and processed through MLP layers. The final set of MLP activations are then concatenated to produce a latent code $\pointdescriptor$.  Formally, this encoder is defined as \eq{local_ndf}.
\begin{align}
    \pointdescriptor = \pointndf(\coord | \pointcloud) = \Phi(\coord | \epsilon(\pointcloud)_{\lfloor \coord \rfloor}).
    \label{eq:local_ndf}
\end{align}
Following \cite{simeonov2021neural}, we utilize occupancy reconstruction to train and learn features for NDFs.

\subsection{Training and Learning SE(3) Equivariance}
\label{sect:equiVar}

To ensure that our models generalize to rigid transformations of the target object, we design a training regime that enforces the descriptors at the same point of an object (in its local frame) to remain invariant under \sethree transformations of the object.

\paragraph{Enforcing SE(3) Equivariance}
In contrast to~\cite{simeonov2021neural}, our system is not inherently SE(3) equivariant.  Instead, we utilize a contrastive loss term to shape the network activations such that they exhibit SE(3) equivariance.  
Formally, an encoder, $\pointndf(\coord | \pointcloud)$, is SE(3) equivariance if, for any rigid body transform $\pose \in \SE3$, 
\begin{equation}
\pointndf(\coord | \pointcloud) \equiv \pointndf(\pose \coord | \pose \pointcloud)
\end{equation}
A simple approach to enforce equivariance is to directly enforce that the encoding of corresponding points should be preserved across SE(3) transformations. However, directly enforcing this constraint was problematic as we found $\pointndf$ to map all inputs to the same encoding.  Therefore, we considered directly enforcing an additional constraint, that different input points produce different encodings, but found the resultant descriptors were no longer semantically consistent between shapes.

\begin{figure}[t]
\centering
\includegraphics[width=0.95\columnwidth]{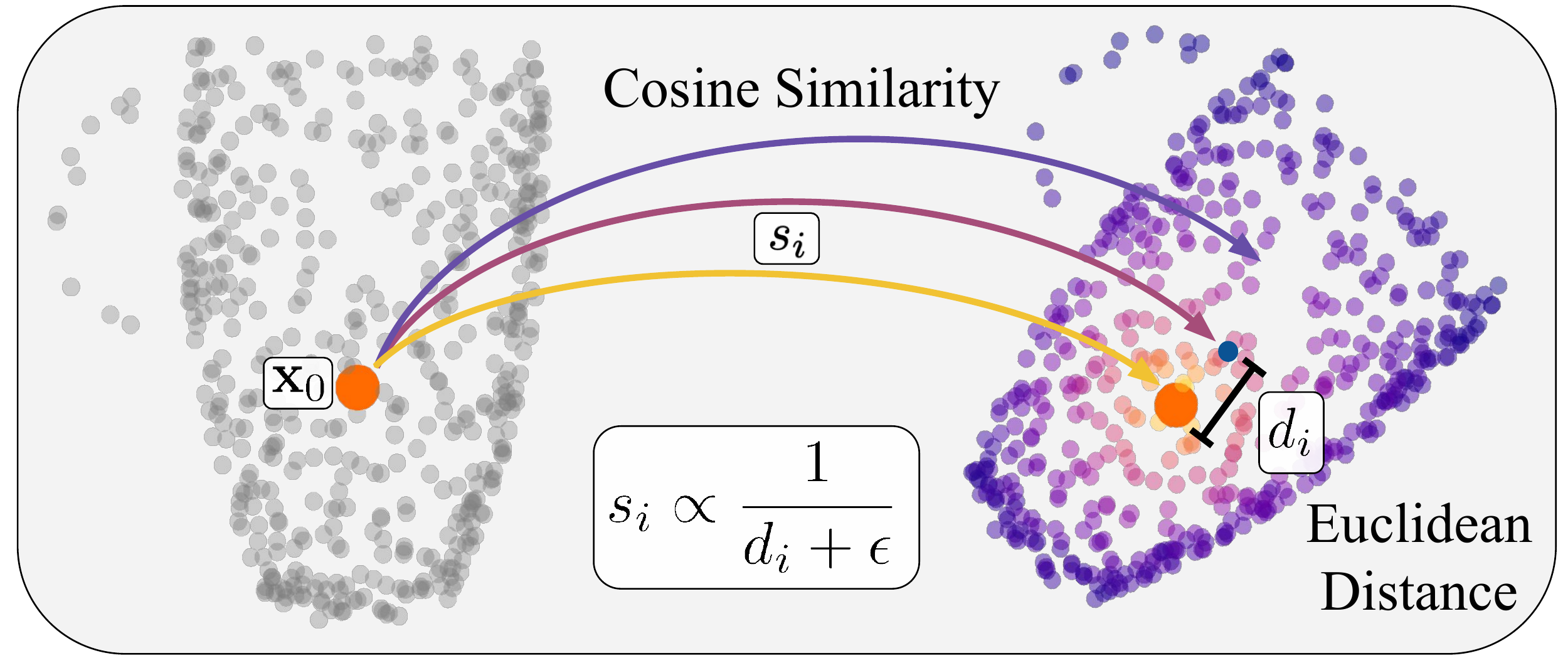}
\caption{\small
\textbf{Contrastive Loss Term for \ourshort} -- The spatial descriptor of a 3D coordinate, $\coord$, with respect to an observed point cloud, $\pointcloud$, is similar across any transform, $\pose \in \SE{3}$.  Additionally, geometrically farther points have decreasingly similar descriptors.
}
\label{fig:loss}
\vspace{-10pt}
\end{figure}

We found that a robust alternative to construct descriptors that are both SE(3) equivariant and semantically consistent was to enforce \eq{inv_dist_sim}, that descriptor similarity between two points is roughly proportional to their inverse distance across different rigid transformations $\pose$ (illustrated in \Fig{loss}). 
\begin{equation}
    \text{sim}(\pointndf(\coord_1|P), \pointndf(\pose \coord_2|\pose P)) \propto \frac{1}{\|\coord_1 - \coord_2\| + \epsilon},
    \label{eq:inv_dist_sim}
\end{equation}
\vspace{0pt}

This constraint enforces that descriptors are both equivariant across rigid transformations of a shape, but also that they vary smoothly with respect to small Euclidean perturbations of the point.

To enforce this loss, we sample $k$ points within the bounding box of the object.  We designate the first point, $\coord_0$, as the point we compute descriptor similarity with respect to in the remaining $k-1$ points.
For each point, we compute the cosine similarity, $s_i$, between $\pointndf(\coord_0 | \pointcloud)$ and $\pointndf(\pose \coord_i | \pose \pointcloud)$,

\begin{align}
    s_i = \frac{\pointndf(\coord_0 | \pointcloud) \cdot \pointndf(\pose\coord_i | \pose\pointcloud)}{\max(||\pointndf(\coord_0 | \pointcloud)|| \cdot ||\pointndf(\pose\coord_i| \pose\pointcloud)||, \epsilon)}
\end{align}
\vspace{0pt}

We compute corresponding target similarity values for each $\coord_i$ with respected to the first point $\coord_0$
\begin{align}
    t_i = \frac{1}{d(\coord_0, \coord_i) + \beta},
\end{align}
\vspace{0pt}

\begin{figure}[t]
\centering
\includegraphics[width=0.95\columnwidth]{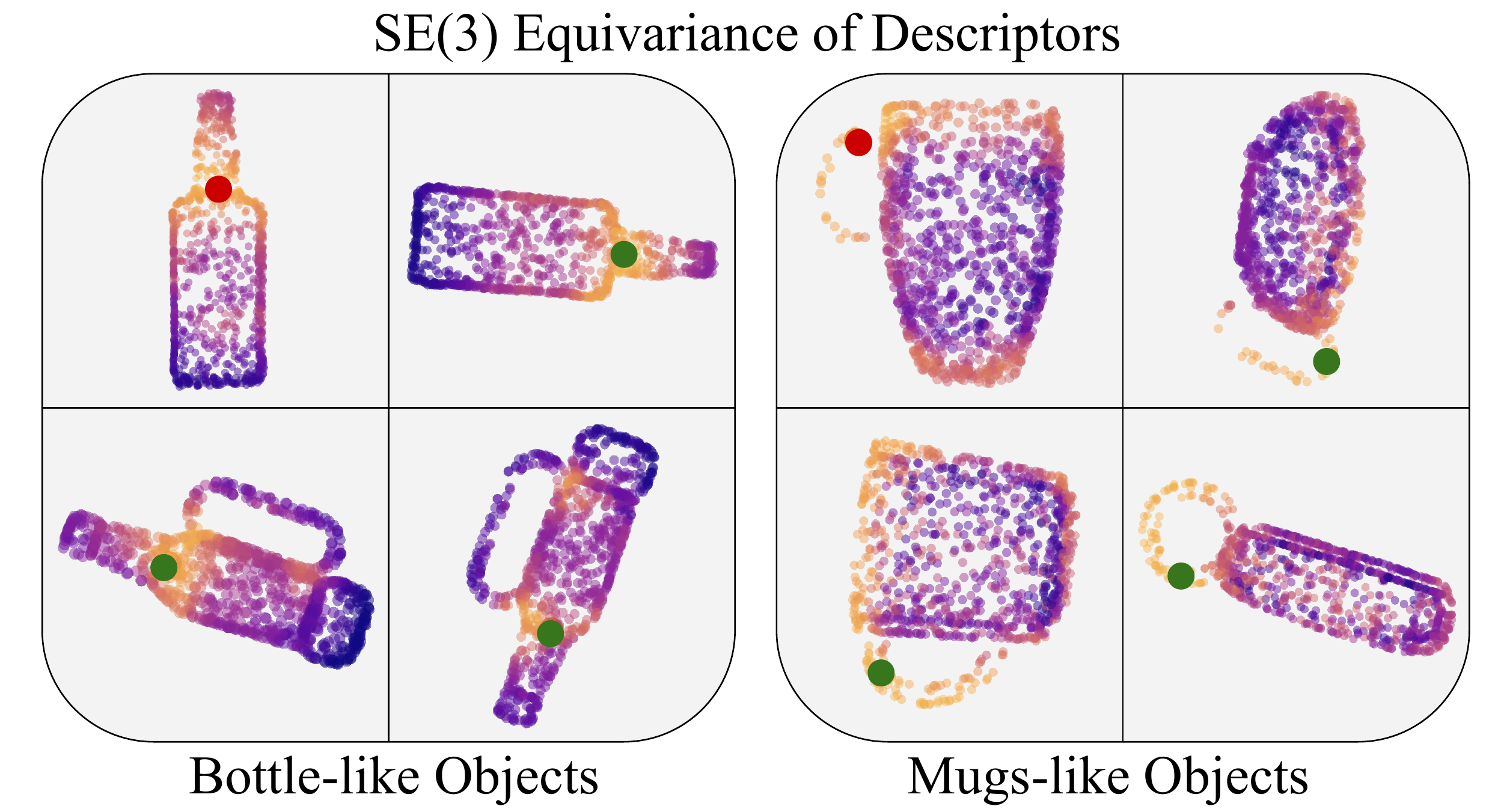}
\caption{\small
\textbf{SE(3) Equivariance of Object Encoding} --
Heat map of cosine descriptor difference from selected point (in red).  The descriptor field remains consistent across different objects in arbitrary SE(3) transformations.
}
\label{fig:se3}
\vspace{-20pt}
\end{figure}

{\setlength{\parindent}{0cm}
and enforce that similarities are roughly proportional to the inverse distance.  As illustrated in \Fig{se3}, this loss successfully enables \sethree equivariance across objects.}

\subsection{Pose optimization}
\label{sect:poseOpt}

When using \ourshort{s} for few shot task learning, we must optimize a pose, $\pose$, on a new point cloud $\pointcloud$, to match a desired reference pose, $\idealpose$ on a reference point cloud $\pointcloud$. This optimization procedure is described in \eq{energy}.  Conventional NDFs run global optimization on a set of query points to obtain the optimal pose $\pose$, where optimization is initialized at a random orientation centered at the origin of the object.  
 
However, this method fails when using Local NDFs.  Since \ourshort{s} only aggregate information across local geometry, there is little information relating distant geometric features.  
To mitigate these challenges, we introduce two techniques: initial translation and query point selection.
 
\paragraph{Initial translation}
In contrast to conventional NDFs, we initialize query points at random rotations and translations within the bounding box of the observed point cloud.  When using a sufficient number of query points instances (We found 20 to be adequate), we find that at least one of the translated query point sets will initialize close to our target geometric feature.  
Subsequent pose optimization tunes the query point cloud to the correct target location.

\begin{figure}[t]
\centering
\includegraphics[width=0.95\columnwidth]{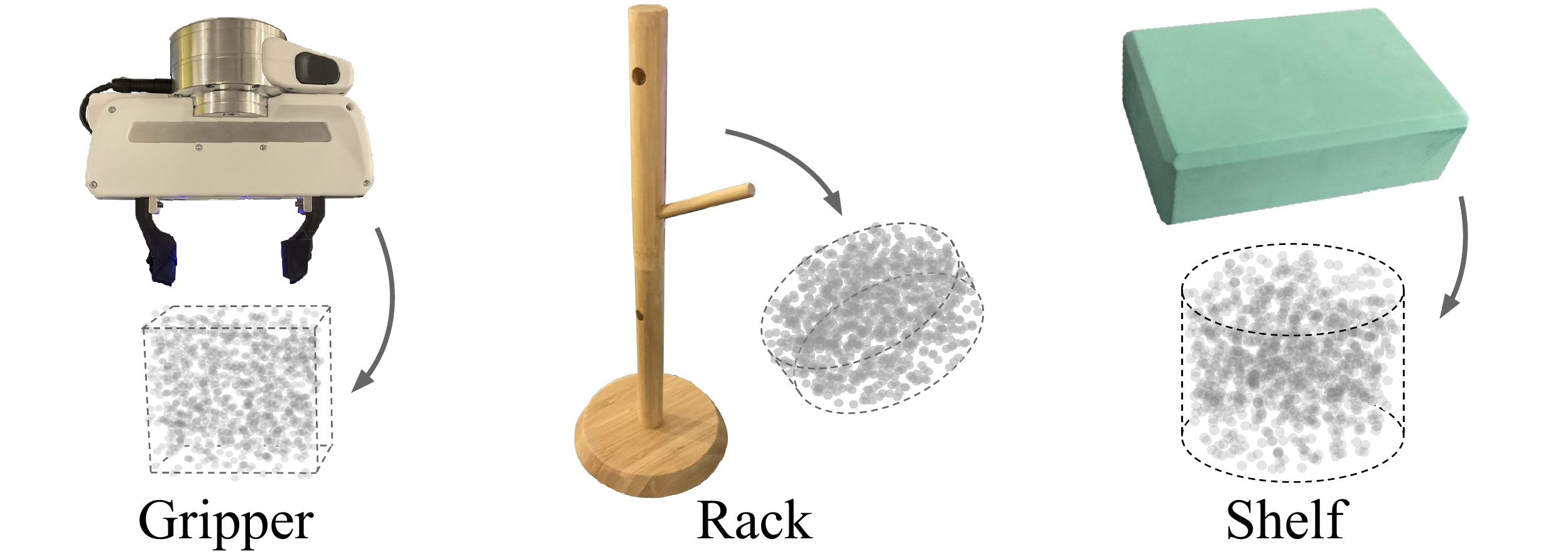}
\caption{\small
\textbf{Selecting Query Points} --
Relative size of query points for each executed task. For grasp and rack placement tasks, we use query points similar in size to contact geometry of the known object (gripper and peg size). For placement surfaces, we find larger query point selections performs well.
}
\label{fig:query}
\vspace{-20pt}
\end{figure}

\paragraph{Query Point Tuning}
We find that query point selection is critical to the performance of \ourshort{s}.  If a query point cloud is too large, it encodes confounding geometry and empty space.  If a query point cloud is too small, it does not capture enough local geometry. 
We find that for precise manipulation, query points can be sampled near the expected contact geometry of the known object.  
For more general poses (such as placing on a surface), a query point cloud which maximizes the expected volume of observed objects contained within the point cloud while minimizing the volume of empty space contained produces robust results.  See \Fig{query} for additional details.

\begin{figure*}[t]

\centering
\captionsetup{type=figure}

\includegraphics[width=\linewidth]{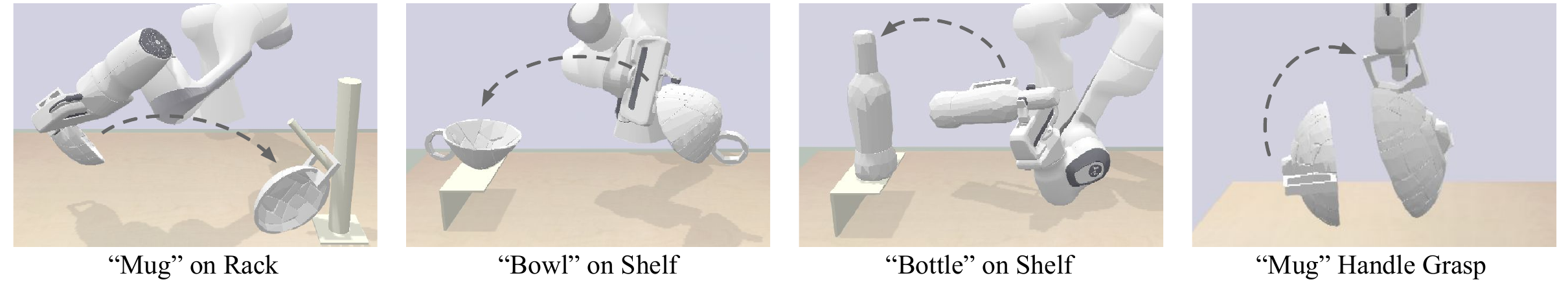}
\captionof{figure}{\small \textbf{Experimental Setup} -- We provide ten simulated demonstrations of each task, then execute each on a set of 200 unseen objects.  We measure grasp success, place success, and overall success. Grasp and place success check that the simulated object is in a stable configuration.  Overall success checks if both grasp and place success occurred.
}

\label{fig:sim_trial}
\vspace{-5pt}
\end{figure*}

\section{Experiments: Design and Setup}

We design our experiments to test the following: (1) How well do \ourshort{'s} generalize to unseen objects classes?  (2) Can \ourshort{'s} be used on a real robot to achieve generalization from a small number of single object class demonstrations?

\subsection{Robot Environment Setup}
Our environment consists of a Franka Panda arm mounted on a table.  Depth cameras are placed at each corner of the table, all calibrated to obtain fused point clouds of objects within the robot’s reach.  We use four depth cameras in simulation, and two depth cameras in real life. 
Our simulation cameras produce a complete point cloud, while the real life cameras produce a partial point cloud.
Depending on the task, a rack or shelf is placed on the table.  For quantitative data, this setup is simulated in Pybullet~\cite{coumans2016pybullet}.  For our simulation setup, refer to \Fig{sim_trial}.  For our real world setup, refer to \Fig{irl_trial}.

\begin{table*}

\definecolor{yellowx}{RGB}{250, 241, 175}
\definecolor{orangex}{RGB}{247, 217, 139}
\definecolor{purplex}{RGB}{253, 189, 255}
\definecolor{bluex}{RGB}{167, 171, 252}

\definecolor{ggreen}{RGB}{220, 240, 213}
\definecolor{gpurple}{RGB}{214, 210 235}
\definecolor{gblue}{RGB}{211, 225, 245}
\definecolor{gred}{RGB}{240, 200, 205}

\newcolumntype{I}{>{\columncolor{ggreen}}X}
\newcolumntype{O}{>{\columncolor{gblue}}X}
\newcolumntype{S}{>{\columncolor{gred}}X}

\footnotesize
\setlength{\tabcolsep}{2.3pt}
\centering
\resizebox{\linewidth}{!}{
\begin{tabularx}{\linewidth}{@{}*{2}lIOSIOOIOIOOS}
    \toprule
        \multicolumn{2}{l}{\textbf{Upright Pose}} & \multicolumn{3}{c}{Mug Demo} & \multicolumn{3}{c}{Bowl Demo} & \multicolumn{2}{c}{Bottle Demo} & \multicolumn{4}{c}{Mug Handle Demo} \\

        & & Mug & Bowl$^*$ & Bottle$^*$
        & Bowl & Bowl$^*$ & Mug 
        & Bottle & Bottle$^*$
        & Mug & Bowl$^*$ & Bottle$^*$ & Bowl\\ 
        
        \cmidrule(lr){3-5} \cmidrule(lr){6-8} \cmidrule(lr){9-10} \cmidrule(lr){11-14}
        
        Geom & Grasp 
        & 0.945 & 0.245 & 0.170
        & 0.670 & 0.605 & 0.305
        & 0.675 & 0.605 
        & 0.350 & 0.375 & 0.510 & 0.215 \\

        & Place
        & 0.360 & 0.455 & 0.330
        & 0.890 & 0.870 & 0.625
        & 0.875 & 0.885
        & - & - & - & -\\
        
        & Overall
        & 0.335 & 0.160 & 0.060
        & 0.590 & 0.530 & 0.205 
        & 0.625 & 0.545
        & 0.350 & 0.375 & 0.510 & \textbf{0.215}\\
    \midrule
        NDF & Grasp 
        & 1.000 & 0.615 & 0.010
        & 0.925 & 0.725 & 0.265 
        & 0.805 & 0.695
        & 0.805 & 0.305 & 0.235 & 0.000\\
        
        & Place
        & 0.925 & 0.620 & 0.225
        & 0.910 & 0.730 & 0.145
        & 0.935 & 0.870
        & - & - & - & -\\
       
        & Overall
        & 0.925 & 0.450 & 0.000 
        & 0.885 & 0.670 & 0.125
        & 0.805 & \textbf{0.665}
        & 0.805 & 0.305& 0.235 & 0.000\\
      
    \midrule
        L-NDF & Grasp 
        & 1.000 & 0.950 & 0.160 
        & 0.990 & 0.985 & 0.970 
        & 0.875 & 0.760 
        & 0.980 & 0.730 & 0.915 & 0.190 \\
        
        & Place
        & 0.995 & 0.830 & 0.900 
        & 0.990 & 0.990 & 0.865 
        & 0.975 & 0.670 
        & - & - & - & - \\
        
        & Overall
        & \textbf{0.995} & \textbf{0.800} & \textbf{0.135} 
        & \textbf{0.985} & \textbf{0.975} & \textbf{0.845} 
        & \textbf{0.850} & 0.590 
        & \textbf{0.980} & \textbf{0.730} & \textbf{0.915} & 0.190 \\
    \midrule
        \multicolumn{2}{l}{\textbf{Arbitrary Pose}} & \multicolumn{3}{c}{Mug Demo} & \multicolumn{3}{c}{Bowl Demo} & \multicolumn{2}{c}{Bottle Demo} & \multicolumn{4}{c}{Mug Handle Demo} \\
      
        & & Mug & Bowl$^*$ & Bottle$^*$
        & Bowl & Bowl$^*$ & Mug 
        & Bottle & Bottle$^*$
        & Mug & Bowl$^*$ & Bottle$^*$ & Bowl\\ 
        
        \cmidrule(lr){3-5} \cmidrule(lr){6-8} \cmidrule(lr){9-10} \cmidrule(lr){11-14}
        
        Geom & Grasp 
        & 0.570 & 0.170 & 0.150
        & 0.730 & 0.690 & 0.555
        & 0.660 & 0.570
        & 0.345 & 0.420 & 0.440 & 0.250 \\
        
        & Place
        & 0.345 & 0.380 & 0.330
        & 0.905 & 0.880 & 0.665
        & 0.850 & 0.860
        & - & - & - & - \\
        
        & Overall
        & 0.215 & 0.075 & 0.065 
        & 0.665 & 0.615 & 0.400
        & 0.600 & 0.525
        & 0.345 & 0.420 & 0.440 & \textbf{0.250} \\
    \midrule
        NDF & Grasp 
        & 0.900 & 0.460 & 0.045 
        & 0.675 & 0.575 & 0.150 
        & 0.575 & 0.385 
        & 0.555 & 0.105 & 0.190 & 0.070 \\
        
        & Place
        & 0.735 & 0.370 & 0.235 
        & 0.840 & 0.800 & 0.565 
        & 0.955 & 0.955 
        & - & - & - & - \\
        
        & Overall
        & 0.655 & 0.250 & 0.010 
        & 0.655 & 0.565 & 0.120 
        & 0.570 & 0.365 
        & 0.555 & 0.105 & 0.190 & 0.070 \\
        
    \midrule
        L-NDF & Grasp 
        & 0.770 & 0.755 & 0.110 
        & 0.910 & 0.960 & 0.880 
        & 0.790 & 0.720 
        & 0.930 & 0.540 & 0.815 & 0.130 \\
        
        & Place
        & 0.960 & 0.635 & 0.850 
        & 0.985 & 0.940 & 0.885 
        & 0.970 & 0.820 &
        - & - & - & - \\
        
        & Overall
        & \textbf{0.735} & \textbf{0.470} & \textbf{0.095} 
        & \textbf{0.905} & \textbf{0.820} & \textbf{0.795} 
        & \textbf{0.775} & \textbf{0.635} 
        & \textbf{0.930} & \textbf{0.540} & \textbf{0.815} & 0.130 \\
    \bottomrule 
\end{tabularx}
\vspace{-5pt}
}

\caption{\small \textbf{Unseen instance pick-and-place success rates in simulation.} Given demonstrations using a single object class, we test performance on a variety of other object classes. NDF performs well on unseen objects from the demonstration object class but struggles with new object classes. L-NDF performs well with unseen objects from both the demonstration and analogous object classes at upright and arbitrary rotations.  An ICP and RANSAC implementation (Geom) is provided as reference.  Green indicates that the test object is the same class as the demonstrations; blue indicates that the test object is from an analogous class to the demonstrations; red indicates that the test object is from a substantially different class.  $^*$Objects are modified to include a handle. See illustrations of each task in \Fig{sim_trial}.}
\label{tbl:benchmark}
\vspace{-15pt}
\end{table*}

\subsection{Task Setup}
We test four tasks: (1) Grasping a mug-like object by its rim and hanging it on a rack by its handle.  (2) Grasping a bowl-like object by its rim and placing it upright on a shelf.  (3) Grasping a bottle-like object by its neck and placing it upright on a shelf.  (4) Grasping a handle placed on an object from an arbitrary object class.  Tasks 1, 2, and 3 use demos containing normal mugs, bowls, and bottles, respectively.  Task 4 uses demos of normal mugs.

We define mug-like objects as standard mugs and bowls with handles attached to them; bowl-like objects as standard bowls, standard mugs, and bowls with handles attached to them; and bottle-like objects as standard bottles and bottles with handles attached to them.

We provide 10 demonstrations per task and test on 200 unseen objects at randomly generated poses, orientations, and uniform scalings.  We assume the environment remains static between demonstrations and test and that (potentially partial) point clouds of the object can be obtained.  In simulation, we use Shapenet~\cite{chang2015shapenet} objects for each in-distribution class, filtering objects that are incompatible with our tasks.  For out-of-distribution objects, we modified Shapenet objects as required. Refer to \Fig{sim_trial} for examples.

\subsection{Training Details}
We pre-train NDFs and \ourshort{'s} by using each system’s occupancy network to reconstruct objects from partial depth images.  We train each system for 300,000 iterations on a joint dataset containing objects from all three object categories at random rotations and translations.  For each object, point cloud data is gathered by placing the object in a PyBullet simulation and taking depth images.  

At test time, we gather a small number (10 in simulation and 4-6 in real life) of task specific demonstrations using a single object class.  These demonstrations are then used by the systems to execute the desired tasks on the demonstration object class, as well as on unseen object classes.

\subsection{Evaluation Metrics}
In simulation, we evaluate each method by measuring grasp success (stable contact between object and end effector), place success (stable contact with placement surface in the correct orientation), and overall task success, for which both grasp success and place success must have occurred.  On the physical robot, human evaluators assert whether the object has been grasped and placed in the correct location.

\subsection{Baselines}
For each of the tasks, we compare \ourshort performance to conventional NDFs and a geometric approach.  The \ourshort query points were selected using the heuristics described above.  NDF query points are extracted from the codebase provided by~\cite{simeonov2021neural}.  The geometric approach uses ICP and RANSAC \cite{Besl_icp_1992, fischler_ransac_1981} for pose estimation.

\section{Experiments: Results}
\label{sec: experiments}

We conduct experiments in simulation to compare the performance of the geometric approach, NDFs, and \ourshort{'s} on each of the four tasks (illustrated in \Fig{sim_trial}) with relevant in and out of distribution objects.  We then perform ablation studies to examine the effect of different loss functions and different 3D feature volumes on L-NDF performance.  Finally, we apply \ourshort{s} on a physical robot and validate that the proposed method generalizes to out-of-distribution poses and objects classes in the real world.

\subsection{Simulation Experiments}
\paragraph{In-distribution objects}
We first consider how skills are transferred to unseen objects from the demonstration class in novel upright or arbitrarily rotated poses.  Referring to the green columns of \Table{benchmark}, we find that in all pick and place tasks, L-NDFs outperform conventional NDFs -- sometimes in excess of a 0.25 increase in success rate.  Furthermore, we find that L-NDFs dramatically outperform NDFs on handle grasping, achieving a 0.38 improvement over NDFs in task success on arbitrarily rotated mug handles (\Table{benchmark}, last green column).  We note that the geometric approach does demonstrate some task succcess; however it lags behind both NDFs and L-NDFs.  We observe that NDF's handle grasp failures occur when a grasp is found near the desired location, but at a slight offset or rotation from the expected location.  Given the fully connected nature of NDFs, we hypothesize that the descriptor fields near an observed object’s salient features may be confounded by the irrelevant geometry of the object itself, an issue which local fields address.

\begin{figure*}[t]

\centering
\captionsetup{type=figure}

\includegraphics[width=\linewidth]{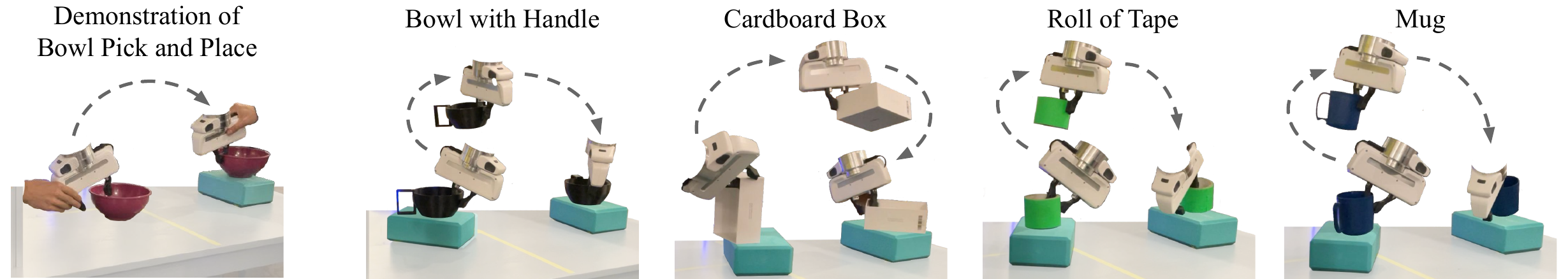}

\captionof{figure}{\small{\textbf{Real world Execution} -- We provide four real world demonstrations of grasping and placing two different bowls.  We then successfully grasp and place a variety of unseen objects using a Franka Panda arm.  Refer to our supplementary video for additional results.}}

\label{fig:irl_trial}

\vspace{-10pt}
\end{figure*}

\paragraph{Analogous Out-of-distribution objects}
We next consider a more difficult task.  We still wish to transfer skills from demonstrations to test objects at novel upright or arbitrarily rotated poses.  However, now the test objects have analogous geometry to the demonstration objects, but in different arrangements or with confounding features. Referring to the first and last blue columns of \Table{benchmark}, we find that on tasks where the rearranged geometry is integral to the task success, NDF performance drops substantially.  In contrast, L-NDF performance does fall, but significantly less than NDF does.  In many cases, we observe that tuning NDF query points to more closely match L-NDF query points can recoup some of this performance loss.  However, we still find that NDF performance lags behind L-NDF success.  

In the middle three blue columns of \Table{benchmark}, we find that, in tasks where the additional feature acts as a confounding feature, NDF and L-NDF overall task success drops by similar amounts. We note that in NDFs, this drop in performance is attributed to both drops in both grasp and placement success.  However, with L-NDFs, this drop is mostly attributed to a decrease in placement success.  
We hypothesize that, while grasping is a highly local task -- only concerned with the location of the manipulator fingers; placement reflects a global task where the orientation of an object is largely defined by its aggregate geometry.  Thus, the advantages of using a local field are diminished in placement and confounding features still affect performance.

\paragraph{Substantially Different Out-of-distribution objects}
Finally, we test the limits of L-NDF's generalization capabilities by testing on objects that are substantially different from the demonstration object class.  Of particular interest is placing a bottle with handle on a rack, given mug demos, and grasping the "handle" of a bowl with no handle (shown in the red columns of \Table{benchmark}). In these extreme cases, we find that NDFs fail completely, achieving negligible success.  L-NDFs fare slightly better, achieving between 10\% and 20\% success.  The geometric approach also shows some success, surpassing \ourshort{s} in the rightmost task.  Interestingly, L-NDFs achieved above 80\% place success on bottles with handles.  As expected, these overall success rates are unsuitable for general robotic manipulation, but suggest that local fields may be a promising direction to explore for more general robotic manipulation.

\subsection{Ablation Analysis}

Next, we run an ablation study on L-NDF using the arbitrary rotation bottle placement task.

\begin{table}[t]
\footnotesize
\setlength{\tabcolsep}{2.3pt}
\centering
\resizebox{\linewidth}{!}{
\begin{tabularx}{\linewidth}{@{}*{12}{X}@{}}
    \toprule
    \multicolumn{3}{c}{Random L-NDF} 
    & \multicolumn{3}{c}{Occupancy Only} 
    & \multicolumn{3}{c}{Hard Contrast}
    & \multicolumn{3}{c}{Distance Contrast} \\
    \cmidrule(lr){1-3} \cmidrule(lr){4-6} \cmidrule(lr){7-9} \cmidrule(lr){10-12}
    G & P & O & 
    G & P & O & 
    G & P & O & 
    G & P & O \\
     \midrule
        0.02 &
        0.73 & 
        0.02 &
        0.70 &
        0.66 &
        0.47 &
        0.64 &
        0.63 &
        0.39 &
        0.79 &
        0.97 &
        0.78 \\
    \bottomrule
\end{tabularx}
} %
\caption{\small \textbf{Effect of Loss Function.}  We test a randomly initialized system and systems trained with pure 3d reconstruction, simple contrastive loss, and our distance based contrastive loss.}  %
\label{tbl:ablation-loss-fn}
\vspace{-10pt}
\end{table}
\begin{table}[t]
\footnotesize
\setlength{\tabcolsep}{2.3pt}
\centering
\resizebox{\linewidth}{!}{
\begin{tabularx}{\linewidth}{@{}*{9}{X}@{}}
    \toprule
     \multicolumn{3}{c}{$32^3$} 
    & \multicolumn{3}{c}{$64^3$} & \multicolumn{3}{c}{$128^3$} \\
    \cmidrule(lr){1-3} \cmidrule(lr){4-6} \cmidrule(lr){7-9}
    Grasp & Place & Overall &
    Grasp & Place & Overall &
    Grasp & Place & Overall \\
     \midrule
        0.63 &
        0.90 &
        0.56 &
        0.77 &
        0.96 &
        0.75 &
        0.79 &
        0.97 &
        0.78 \\
    \bottomrule
\end{tabularx}
} %
\caption{\small \textbf{Effect of 3D Feature Volume Size.} 
We examine the effect of 3D feature volume size (in voxels) on L-NDF performance.  All systems are trained using our distance based contrastive loss}  %
\label{tbl:ablation-voxel}
\vspace{-15pt}
\end{table}

\paragraph{Loss Function}
First, we analyze the impact of the loss function on L-NDF performance. In \Table{ablation-loss-fn}, we find that a random network achieves negligible grasp success and subpar place success. This indicates that pretraining L-NDF is important.
A simple contrastive loss function where similar points have ground truth similarity of 1 and different points have ground truth similarity of 0 performs poorly as well.  We hypothesize that enforcing this sort of loss incorrectly describes our objectives for the network, as different example points should, intuitively, have different costs.  Solely training on reconstructive tasks performs better than simple contrastive loss, but yields poor performance at arbitrary rotations.  However, our distance based contrastive loss dramatically improves on both methods, enforcing SE(3) equivariance while preserving reconstruction quality.

\paragraph{3D Feature Volume Size}
We next analyze the impact of voxel size on L-NDF performance. Referring to \Table{ablation-voxel}, we find that task success monotonically increases with 3D feature volume size.  Increasing the feature volume from $32^3$ voxels to $64^3$ voxels produces a dramatic improvement, while increasing from $64^3$ voxels to $128^3$ voxels produces increased success, but at a diminishing rate.  We elect to use the $128^3$ voxel system as it ran in similar time to the $64^3$ voxel while providing slightly higher success rates.

\subsection{Real world}

\begin{figure}[t]
\centering
\includegraphics[width=0.95\columnwidth]{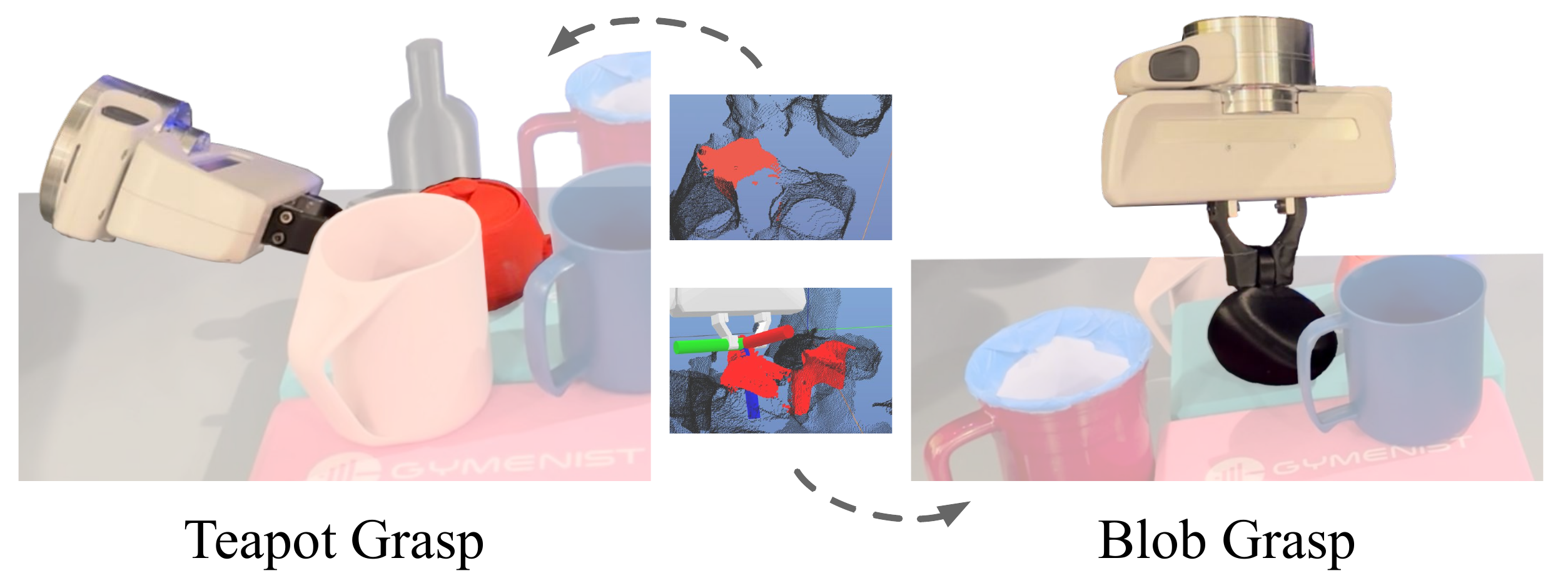}
\caption{\small
\textbf{Operating in Clutter} --
We provide four real world demos of grasping a mug in an uncluttered scene.  We then grasp a teapot and blob in a cluttered environment using partial point clouds.  We used Mask R-CNN \cite{he_maskrcnn} for scene segmentation.  Please see our supplementary video for additional results.
}
\label{fig:clutter}
\vspace{-20pt}
\end{figure}

Finally, we evaluate our system in a real world environment.  We collect 5-10 task demonstrations for handle grasping and bowl pick and place using upright objects, then evaluate our system on a variety of unseen objects in arbitrary poses.  Additionally, we evaluate our system in a cluttered environment, using Mask R-CNN \cite{he_maskrcnn} for scene segmentation and L-NDF for pose estimation.  As can be seen in \Fig{clutter}, the resultant point clouds from scene segmentation are often incomplete and noisy, yet LNDF successfully deduces object pose.  Please see \Fig{teaser} and \Fig{irl_trial} for our single object trials, \Fig{clutter} for our evaluation in cluttered environments, and our website for additional details and qualitative results.

\section{Conclusion}
\label{sec:conclusion}

We introduce \ours{}s, an object representation that allow few-shot imitation learning of manipulation tasks on potentially novel categories of shapes at test time. We illustrate the capability of our work to exhibit strong generalization -- given only examples of grasping the handle of a mug, we can generalize to shapes such as teacups or bottles in both simulation and the real world.

\section{Acknowledgement}
\label{sec: ack}

We gratefully acknowledge support from NSF grant 2214177; from AFOSR grant FA9550-22-1-0249; from ONR MURI grant N00014-22-1-2740; from ARO grant W911NF-23-1-0034; from the MIT-IBM Watson Lab; and from the MIT Quest for Intelligence.

\renewcommand*{\bibfont}{\normalsize}
\begin{flushright}
\printbibliography %
\end{flushright}

\end{document}